\documentclass{article}

\usepackage[nonatbib,preprint]{nips_2018}

\usepackage[utf8]{inputenc} 
\usepackage[T1]{fontenc}    
\usepackage{hyperref}       
\usepackage{url}            
\usepackage{booktabs}       
\usepackage{amsfonts}       
\usepackage{nicefrac}       
\usepackage{microtype}      
\usepackage{amsmath}
\usepackage{amssymb}
\usepackage{xcolor}
\usepackage{subcaption}
\usepackage{graphicx}
\usepackage{algorithm,algpseudocode}
\usepackage{coolMath}
\usepackage{slashbox}
\usepackage{multirow}
\usepackage{array}

\newcolumntype{P}[1]{>{\centering\arraybackslash}p{#1}}
\newcommand\T{\rule{0pt}{2.9ex}}       

\title{Neural Generative Models for Global Optimization with Gradients}

\author{
  Louis Faury \thanks{Corresponding author.}\\
  Criteo Research\\
  France\\
  \texttt{l.faury@criteo.com} \\
  \And
  Flavian Vasile\\
  Criteo Research\\
  France\\
  \texttt{f.vasile@criteo.com}
  \And 
  Cl\'ement Calauz\`enes\\
  Criteo Research\\
  France\\
  \texttt{c.calauzenes@criteo.com}
  \And 
  Olivier Fercoq\\
  LTCI,
  T\'el\'ecom ParisTech\\
  Université Paris-Saclay, France \\
  \texttt{olivier.fercoq@telecom-paristech.fr}
  }

\begin{document}

\maketitle

\begin{abstract}
 	The aim of global optimization is to find the global optimum of arbitrary classes of functions, possibly highly multimodal ones. In this paper we focus on the subproblem of global optimization for differentiable functions and we propose an Evolutionary Search-inspired solution where we model point search distributions via Generative Neural Networks. This approach enables us to model diverse and complex search distributions based on which we can efficiently explore complicated objective landscapes. In our experiments we show the practical superiority of our algorithm versus classical Evolutionary Search and gradient-based solutions on a benchmark set of multimodal functions, and demonstrate how it can be used to accelerate Bayesian Optimization with Gaussian Processes.
\end{abstract}

\section{Introduction}
{
	In the sphere of practical applications for global optimization, there are many types of problems in which the cost of evaluating the objective function is high, rendering brute force methods inefficient. Two classical examples of such problems are \emph{hyper-parameters search} in machine learning and \emph{model fitting} - which often requires running long and expensive simulations. 
    In many cases, derivatives of the objective function are not accessible or too expensive to compute, restricting the set of solvers to zero order algorithms (using function evaluations only).
    Still, global optimization with gradients covers a great diversity of interesting problems, like the efficient maximization of acquisition functions in Bayesian Optimization (BO) \cite{jones2001taxonomy} or the training of machine learning models with non-decomposable objectives such as Sample-Variance Penalization \cite{Maurer2009,Swaminathana} -- decomposable ones being commonly optimized using stochastic methods.  Recent work \cite{pedregosa2016hyperparameter, maclaurin2015gradient} also opened the door to affordable computations of hyper-parameters derivatives, pushing for the development of efficient first order methods for global optimization.

One popular class of zero order global optimization algorithms is Evolutionary Search (ES). ES methods have recently seen a growing interest in the machine learning community, namely for the good results of Covariance Matrix Adaptation Evolutionary Strategies (CMA-ES) \cite{hansen2001completely} for hyper-parameter search and Natural Evolution Strategies (NES) \cite{wierstra2008natural, salimans2017evolution} for policy search in reinforcement learning. 
    The main idea behind ES is to maintain a population of points where to evaluate the objective function $f$. Points with small objective value are kept for the next population (\emph{selection}), combined together (\emph{recombination}) and slightly randomly modified (\emph{mutation}). This process is repeated iteratively, and relies on the internal noise to explore $f$'s landscape. In both CMA-ES and NES, populations are sampled from Gaussian distributions, their moments being modified iteratively along the optimization procedure. 
    
    If the simplicity of Gaussian search distributions allows computational tractability, it can also slow down the optimization process. For complicated functions, ellipsoids can give a poor fit of the areas with small objective values, which hinders the exploration process. In Figure \ref{fig::cmarosenbrock} we show the evolution of $p_\theta$ under the CMA-ES algorithm for the Rosenbrock function. With the provided initialization, the Gaussian search distribution doesn't perform efficient exploration as it cannot adopt the curvature imposed by the landscape. It expands and retracts until it finally fits into the straight part of the curved valley. Using more diverse and general search distributions is therefore an interesting way to improve evolutionary search. One of the most promising class of models for fitting search distributions is the class of generative neural networks \cite{dziugaite2015training,mackay1995bayesian} that have been shown to be able to model non-trivial multimodal distributions \cite{cao2018improving, vukotic2017generative}.
    
    \begin{figure}[h!]
    	\centering
        \begin{subfigure}[b]{0.33\linewidth}
			{
				\centering
				\includegraphics[width=\textwidth]{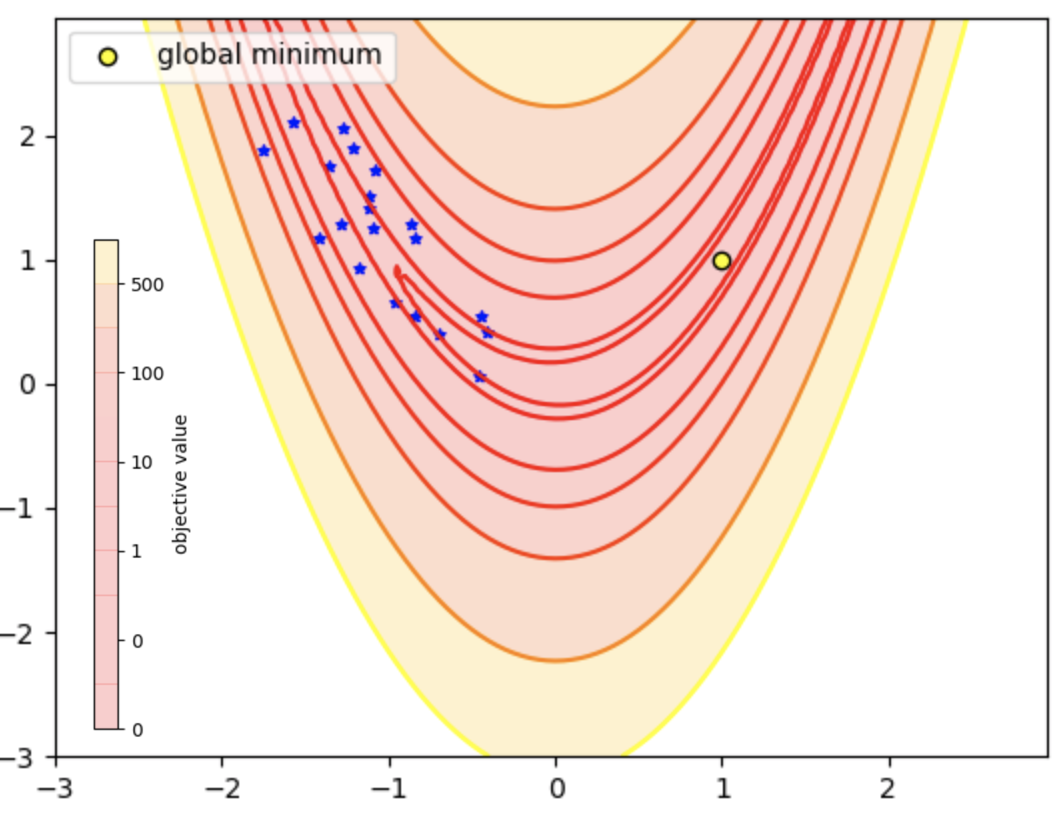}
				\caption{Iteration 0}
			}
    \end{subfigure}\hfill
    \begin{subfigure}[b]{0.33\linewidth}
			{
				\centering				\includegraphics[width=\textwidth]{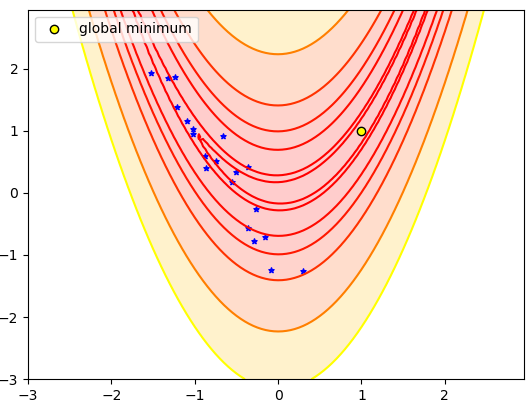}
				\caption{Iteration 1}
			}
    \end{subfigure}\hfill
    \begin{subfigure}[b]{0.33\linewidth}
			{
				\centering
				\includegraphics[width=\textwidth]{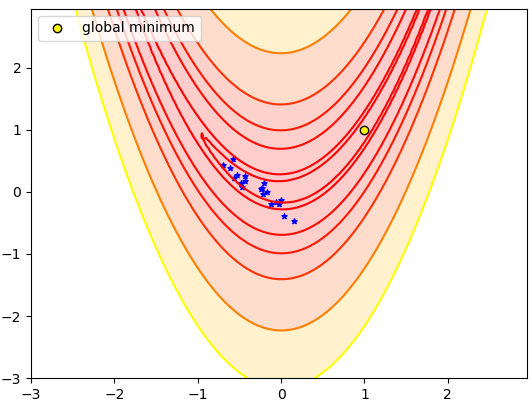}
				\caption{Iteration 10}
			}
    \end{subfigure}
    
    \caption{Evolution of samples from $p_\theta$ on the Rosenbrock function for CMA-ES.}
    \label{fig::cmarosenbrock}
    \end{figure}
        
    Our main contribution is to replace Gaussian search distributions by generative neural networks, that allows us to use gradient information and to improve both convergence speed and quality of found minima. Along with introducing this new type of architecture we show how to train it efficiently, and how it can be used to quickly optimize multimodal differentiable functions. Interestingly enough, \cite{lopez2018easing} recently proposed a fairly similar method, though motivated by over-parametrization and not by an ES approach.
    The rest of this paper is organized as follows: in Section \ref{sec::pb}, we present related work on global optimization with zero and first order oracles. We describe in Section \ref{sec::gennes} the algorithm we propose, discuss its links with ES and provide insights into its architecture settings. In Section \ref{sec::experiments} we report experiments on a continuous global optimization benchmark, and show how our algorithm can be used to accelerate BO. We conclude and discuss future work in Section \ref{sec::conclusion}. 
}

\section{Problem formulation and related work}
{
	\label{sec::pb}
    In global optimization the task is to find the global optimum of function $f$ over a compact set $\mathcal{X}$:
    \begin{align*}
    x^* = \argmin_{x \in \cX} f(x)
    \end{align*}
    For general classes of functions, this cannot be performed greedily and requires the exploration of the associated landscape. Below, we briefly present zero and first order algorithms designed at solving this problem.
    
    \subsection{Zero order algorithms} When only provided with a zero order oracle of the objective, state-of-art methods are either based on evolutionary search or on designing a surrogate of the objective integrating optimism to ensure exploration - which is at the heart of the BO framework.
    
    \paragraph{Evolutionary Search}
    A thorough overview of ES is out of the scope of this paper. Here we briefly present the NES class of algorithms and CMA-ES, for they provide intuition for our method. Both these methods maintain a set of points (called population) sampled from a parametric search distribution $p_\theta$, at which they evaluate the objective $f$. From these function evaluations, NES produces a search gradient on the parameters towards lower expected objective. Formally, it minimizes: 
		\begin{equation}
			J(\theta) = \mathbb{E}_{x\sim p_{\theta}}\left[ f(x)\right]
            \label{eq::obj}
		\end{equation}
        encouraging $p_{\theta}$ to steer its probability mass in area of smallest objective value. One version of the NES uses the score function estimator for $\nabla_\theta J$ using the likelihood-ratio trick in close resemblance to \cite{williams1992simple}: 
		\begin{equation}
				\nabla_\theta J(\theta) =  \mathbb{E}_{x\sim p_\theta}\left[f(x)\nabla_\theta\log p_\theta(x)\right]
		\end{equation}
		that can be approximated from samples, and then feeds it into a stochastic gradient descent algorithm to optimize $\theta$. Note that this requires a closed form for $p_\theta$, which is chosen to be a Gaussian distribution for its tractability, with $\theta$ describing its mean and covariance matrix. Although it doesn't explicitly minimizes \eqref{eq::obj}, CMA-ES aims at the same objective. The search distribution $p_\theta$ is tied to be Gaussian and relies on Covariance Matrix Adaptation for the update. Formally (for its simplest form), CMA-ES retains a fraction of the best points within the population that are used to update the mean and variance of $p_\theta$. We refer the interested reader to \cite{hansen2016cma} for thorough details on the procedure followed by CMA-ES.

    \paragraph{Bayesian Optimization} Bayesian Optimization (BO) is a popular machine-learning framework for global optimization of expensive black-box functions that doesn't require derivatives. The main idea behind BO is to sequentially: (a) fit a model of the objective given the current history of function evaluations $\mathcal{H}_t:=\left\{ x_{1:t},f(x)_{1:t}\right\}$, (b) use this model to identify a promising future query $x^*_t$, (c) sample $f(x^*_t)$ and add it to $\mathcal{H}$. BO has shown state-of-the-art performances for model fitting \cite{acerbi2017practical} or hyper-parameter tuning in machine learning \cite{snoek2012practical}.
    
        Gaussian Processes (GPs) provide an elegant way to model the function provided $\mathcal{H}$ and to balance the exploration/exploitation trade-off in a principled manner, by defining priors over functions. Let $k$ be a positive definite kernel over $\mathcal{X}$. The $\text{GP}(0,k)$ prior imposes that for any collection $\mathbf{x}=(x_1,\hdots,x_t)^T$, the vector $\mathbf{f}=(f(x_1),\hdots,f(x_t))^T$ is a multivariate Gaussian with mean $0$ and covariance $K=(k(x_i,x_j))_{i,j}$. Using conditional rules for Gaussian distribution, one can show that $p(f\vert \mathcal{H})$ is also a Gaussian, whose mean and covariance are:
        \begin{align}
        	m(x) &= k(x,\mathbf{x})K^{-1}\mathbf{f}\\
            \sigma^2(x) &= k(x,x)-k(x,\mathbf{x})K^{-1}k(\mathbf{x},x)
        \end{align}
        These derivations as well as many details and insights on GPs can be found in \cite{rasmussen2004gaussian}.
The exploitation/exploration trade-off is ruled by the acquisition function, which role is to find the point $x^*_t$ where the improvement over the current minimum $\tilde{f}=\min_{1\leq i \leq t} f(x_i)$ is likely to be the highest. In this work, we focus on one popular acquisition function: the Expected Improvement (EI, \cite{jones2001taxonomy}), although there exists many others (\cite{villemonteix2009informational} ,\cite{srinivas2009gaussian}) which could be equivalently used in this work. After fitting a GP on $\mathcal{H}$, we obtain a distribution $p_x\triangleq\mathcal{N}\left(m(x),\sigma^2(x)\right)$ for the value of the objective at every test point $x\in\mathcal{X}$. The EI is defined as:
            \begin{equation}
            \label{eq:def_EI}
            	\text{EI}(x) = \mathbb{E}_{y\sim p_x}\left[ \max{(\tilde{f}-y,0)}\right]
            \end{equation}
            The EI is non-negative and takes the value $0$ at every of the training points. It is likely to be highly multimodal, especially when there is a large number of training points. It is generally smooth as it inherits the smoothness of the kernel $k$ (usually at least once differentiable). It could seem that so far, we just transposed the problem of minimizing the objective to the maximization of the acquisition function to find $x_t^*$, where the expected improvement is the highest. However, in the BO setting, a common assumption is that the acquisition function is excessively cheaper to evaluate than the actual objective, which allows for brute force approach of its optimization. Still, GP inference cost grows cubically with the number of queried points and therefore in practice, BO becomes inefficient in high dimensions when one would need a large number of points before locating a global optimum. Being able to reduce the number of acquisition function evaluations is crucial for scaling BO. Also, note that the acquisition function has computable derivatives which might be very useful for its fast global maximization. 
            

    \subsection{First-order algorithms} Surprisingly, there exist very few gradient-based methods for global optimization. A popular technique is to repeat greedy algorithms like gradient descent from a collection of initial starting points. This method has become very popular in the BO community, where repeated BFGS has become the standard way to optimize the acquisition function, as it (almost always) has computable derivatives. The initial starting points are usually sampled uniformly over the compact set $\mathcal{X}$, or according to adaptive grid strategies. Note that this initial population could also be the result of an ES procedure, although we are not aware of any work implementing this idea. 

Recently, it was proposed to incorporate derivative information in the BO framework, either to improve the conditioning of the GP covariance matrix \cite{osborne2009gaussian}, or to affect directly the acquisition function \cite{lizotte2008practical,wu2017bayesian}. However, the previous remarks on BO scalability still apply to these approaches.
}

\section{Generative Neural Network for Differentiable Evolutionary Search}
{
    \label{sec::gennes}
    NES and CMA-ES choose the search distribution $p_\theta$ to be Gaussian, although any parametric distribution could be used to generate populations. A general way to construct one is to apply a parametric transformation to an initial random variable $u$, which probability distribution we note $\mathcal{P}$. In our case, the parameter vector $\theta$ of this transformation has to be adapted or learned so that $p_\theta$ can be used to explicitly optimize $f$. 
    
    Neural networks are able to generate complex transformations and their weights and biases can be learned quickly thanks to gradient back-propagation, and therefore they constitute good candidates for generating $p_\theta$ from $\mathcal{P}$. We here describe the method we propose to leverage this idea, called \emph{Generative Neural Network for Differentiable ES - GENNES}.
    
    \subsection{Core algorithm}
    {
	We note $G_\theta$ the neural network parametrized by $\theta$ (the weights and biases of the network), mapping the noise $u\sim \mathcal{P}$ into points $x \in\mathcal{X}$. As  our goal is to generate queries $x$ with low-value of the objective, \eqref{eq::obj} is a natural cost function for training $G_\theta$. However, note that we don't have access to a closed form of $p_\theta$ and therefore ideas similar to NES cannot be applied. Still, \eqref{eq::obj} can be rewritten as:
	\begin{equation}
		J(\theta) = \mathbb{E}_{u\sim \mathcal{P}}\left[f(x(\theta,u))\right]
	\end{equation}
	where $x(\theta,u)$ is the output of $G_\theta$ with input $u$. This allows us to compute a stochastic estimate of $J$'s gradient with respect to $\theta$:
	\begin{equation}
		\begin{aligned}
			\nabla_\theta J(\theta) &\simeq \frac{1}{N} \sum_{i=1}^N \frac{\partial}{\partial{\theta}}f(x(\theta,u_i))\\
							   &= \frac{1}{N} \sum_{i=1}^N \frac{\partial x}{\partial \theta}(\theta,u_i)^T\nabla_x f(x(\theta,u_i))
		\end{aligned} 
	\end{equation}
	where $\{u_1,\hdots,u_N\}$ is a collection of samples from $\mathcal{P}$. This stochastic estimate of the gradient is then fed to a stochastic gradient descent algorithm like Adam \cite{kingma2014adam}. Note that the Jacobian term $\frac{\partial x}{\partial \theta}$ of the neural network can be easily computed via back-propagation. Figure \ref{fig::gennespop} demonstrates the behavior of a naive implementation of GENNES on the Rosenbrock (unimodal, poorly conditioned) function and the Rastrigin function (highly multimodal). The batch size $N=40$ used here for the sake of illustration is purposely much larger than it would practically be for such low-dimensional problems to better describe the support of $p_\theta$. It is noticeable that GENNES is able to learn a curved-shaped search distribution on the Rosenbrock, in accordance to the contour lines of the function. On Rastrigin's function, we see that GENNES is able to explore several minima (including the global one) at the same time. This is very useful for multimodal objectives, where the exploration has to be conducted in disconnected areas of the landscape. We provide in Algorithm \ref{algo::gennes} the pseudo-code for GENNES.
    
    \begin{algorithm}[h]
				\caption{Generative Neural Model for Differentiable ES (GENNES)}
\label{algo::gennes}			\begin{algorithmic}[1]
					\Procedure{GENNES}{$f,\mathcal{P},\theta_0,N,T,\eta$}
					\State Initialize $G_{\theta}$ with $\theta_0$.
   					\For{$t\leftarrow 1,T$}
						\State Sample $(u_1^t,\hdots,u_N^t)\sim P$ i.i.d. 
						\State Generate $(x_1^t,\hdots,x_N^t) := (x(u_1^t,\theta),\hdots,x(u_N^t,\theta))$ by a feed-forward pass of $G_\theta$. 
						\State Query $f$ and $\nabla_xf$ at $x_i^t$, $i=1,\hdots,N$.
                    \State Compute the stochastic estimator:
                    $$
                    	\Delta J_t = \frac{1}{N}\sum_{i=1}^N \frac{\partial x}{\partial \theta}(u^t_i,\theta)^T \nabla_x f(x(u_i^t,\theta))
                    $$
                    \State Apply Adam with learning rate $\eta$ and gradient estimate $\Delta J_t$ to $G_\theta$. 
					\EndFor \\
                    \Return $\text{argmin}_{x_i^t} f(x_i^t)$
					\EndProcedure
				\end{algorithmic}
			\end{algorithm}
    
    \begin{figure}
		\centering
		\begin{subfigure}[b]{0.33\linewidth}
			{
				\centering
				\includegraphics[width=\textwidth]{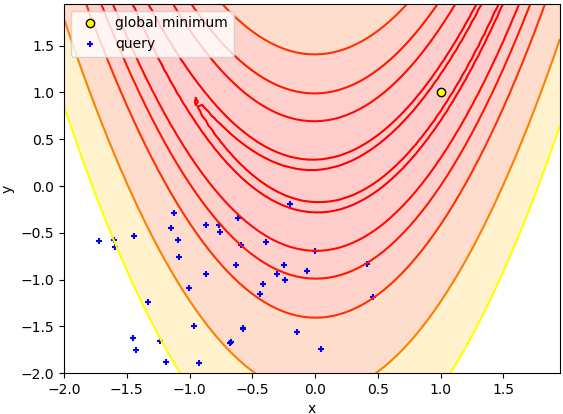}
				\caption{Iteration 0}
			}
			\end{subfigure}\hfill
			\begin{subfigure}[b]{0.33\linewidth}
			{
				\centering
				\includegraphics[width=\textwidth]{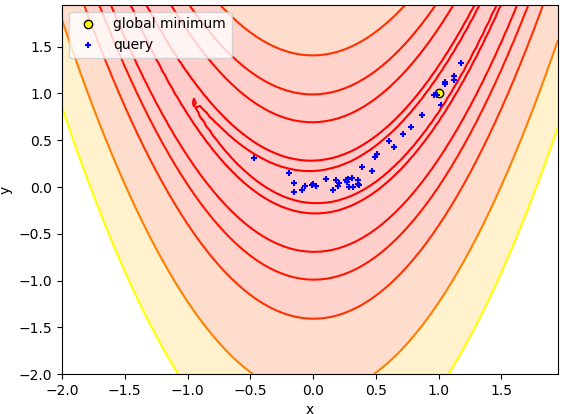}
				\caption{Iteration 10}
			}
			\end{subfigure}\hfill
			\begin{subfigure}[b]{0.33\linewidth}
			{
				\centering
				\includegraphics[width=\textwidth]{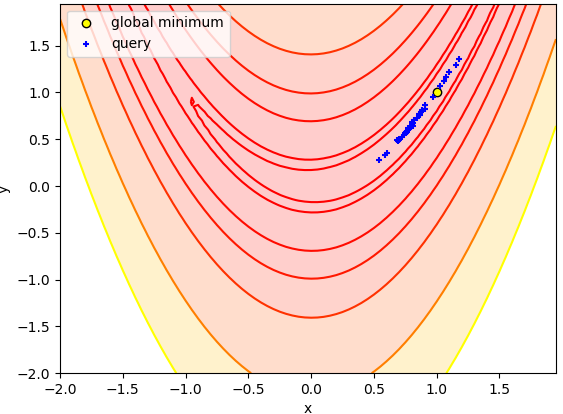}
				\caption{Iteration 50}
			}
			\end{subfigure}\\
        \addtocounter{subfigure}{-3}
		\begin{subfigure}[b]{0.33\linewidth}
			{
				\centering
				\includegraphics[width=\textwidth]{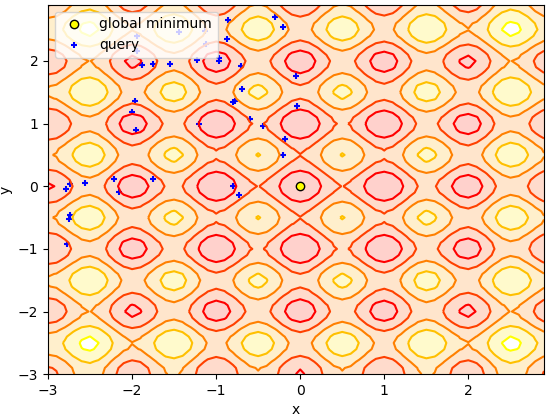}
				\caption{Iteration 0}
			}
			\end{subfigure}\hfill
			\begin{subfigure}[b]{0.33\linewidth}
			{
				\centering
				\includegraphics[width=\textwidth]{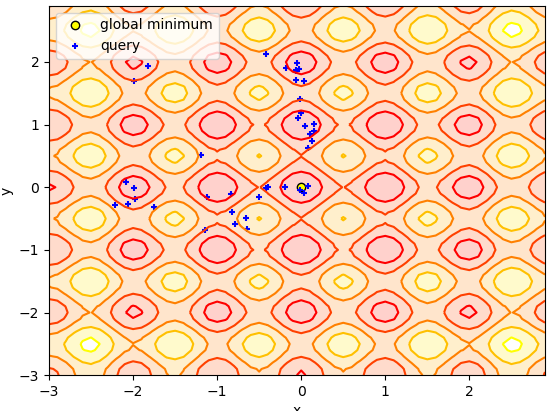}
				\caption{Iteration 20}
			}
			\end{subfigure}\hfill
			\begin{subfigure}[b]{0.33\linewidth}
			{
				\centering
				\includegraphics[width=\textwidth]{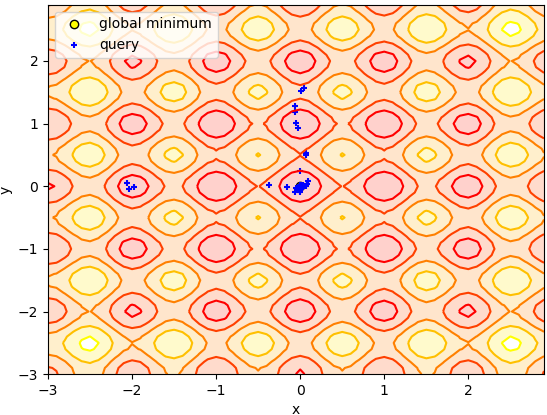}
				\caption{Iteration 100}
			}
			\end{subfigure}\hfill
			\caption{Evolution of $p_\theta$ with GENNES on the Rosenbrock (top) and Rastrigin (bottom) functions.}
			\label{fig::gennespop}
	\end{figure}

    \subsection{Links with ES}
    GENNES shares many intuitive links with ES. For instance, the batch size $N$ can be interpreted as a population size. Indeed, the batch $\{x(\theta,u_1),\hdots,x(\theta,u_N)\}$ can be understood as a population, which fitness signals (\emph{e.g} $f$'s gradients) are used to encourage $G_\theta$ into generating better individuals at the next generation - \emph{i.e} next update. In short, $G_\theta$ pushes search points $x$ along $f$'s negative gradients. However, because its capacity is limited and all the search points are generated by the same model, it is forced to mix and balance information coming from different queries, which can be understood as the recombination process of ES. Sampling different noise $u$ at different iterations can itself be understood as a mutation step. We optimize $G_\theta$ with Adam, as we noticed training the network with momentum gives inertia to the output distribution, leading to faster optimization. From an ES perspective this can be understood as using an \emph{evolution path} \cite{hansen2015evolution}. 

    \subsection{Architecture and refinements}
    {
    	
        \paragraph{Architecture} In all our experiments we use fully connected layers with leaky ReLU \cite{maas2013rectifier} activations  to map the input noise $u$ to an intermediate embedding layer. We then add a fully connected layer with hyperbolic tangent activations as the final layer of the generator $G_\theta$. This allows us to impose the output distribution to lay on a compact set - as we usually set bounds to the domain of interest over which we want to optimize the objective $f$. We don't use any renormalization techniques like batch normalization or weight normalization as we found that a simple architecture performed best in our experiments. The weights are initialized with Glorot initialization \cite{glorot2010understanding}, expect for those belonging to the last layer which are initialized according to a normal distribution. The noise is sampled according to a multivariate uniform distribution. The dimension, as well as the support of this distribution, are important hyper-parameters as they shape the form of the initial output distribution. An inconvenient shape (due to the noise dimension being way smaller than the objective dimension) or a degenerate support (if $u$'s standard deviation is too small) can seriously hinder the optimization process. To this end, we provide in Appendix \ref{sec::safeinit} a robust adaptive initialization strategy to set the variance of the initial weights of the final layer and $u$'s magnitude.

        \paragraph{Noise annealing} An interesting problem we encountered was the one of precision. With the architecture we described until here, \textit{GENNES} is able to quickly locate $f$'s global minimum but fails to sample points arbitrarily close to it, leading the optimization procedure into what appears as a premature convergence. Ideally, we want $p_\theta$ to converge to a Dirac located at the global minimum, which means that $G_\theta$ should map the whole support of $u$ into a single point $x^*$. This either requires infinite capacity of the network, or to \emph{unlearn} the input - \emph{i.e} set all weights of the first layer to zero. Both solutions being unrealistic, we propose to progressively reduce the support of $u$. This procedure, which we refer to as \emph{noise annealing} in similarity to the simulated annealing algorithm, can be used if high precision on the global minimum is required. Noise annealing has an interesting effect of the optimization procedure: as the support of the input distribution shrinks, so does the support of the output distribution, forcing it to be unimodal and with small support which progressively makes it follow the negative gradient of the function. To some extent, we can consider it as applying gradient descent to the most fit particle after a first evolutionary search procedure. Rules of thumbs are given in Appendix \ref{sec::noiseschedule} for setting the schedule of the noise annealing.  
    }
}

\section{Experimental Results}
{
	\label{sec::experiments}
	In this section, we first evaluate GENNES on a small continuous global optimization benchmark. The good results we obtain motivate the second set of experiments, where we show how GENNES can accelerate the BO procedure on a toy example and a hyper-parameter search experiment. 
    
	\subsection{Continuous optimization benchmark}
    {
    	\label{sec::continuousopt}
    	\paragraph{Experimental set-up} Here, we describe the experimental results obtained on four functions taken from the noiseless continuous optimization benchmark BBOB-2009 \cite{hansen:inria-00362633} testbed: Rastrigin, Ackley, Styblinksi and Schwevel functions, which literal expressions can be found in Appendix \ref{sec::functionformulas}. All these functions are multimodal, although Ackley has much more local minima than others on the domain it is evaluated on. It also has the best global structure as the global minimum is at the center of a globally decreasing landscape.
For every problem instance, we repeat the optimization procedure for a 10-fold, and report mean regret as a function of number of evaluations. At every new fold, the position of the global minimum of the objective is randomly translated, and the weights of the generator for GENNES are randomly initialized with a different seed.

\paragraph{Baselines} We compare our algorithm to a repeated version of L-BFGS (\emph{i.e} ran from multiple start points), and also provide the results obtained by the respective authors implementations of aCMA-ES \cite{hansen2010benchmarking} and sNES \cite{schaul2012benchmarking}. Both these algorithms use zero-order oracle for optimization, while our algorithm and L-BFGS use first-order oracle. They however provide state-of-the art results for global optimization of such functions and therefore allow to discuss the validity of the minima discovered by GENNES. It is important to keep in mind that we count a gradient evaluation as equal to a function evaluation of the objective, therefore conclusions of these experiments will only hold in settings where this rule is reasonable. We run aCMA-ES and sNES with default (adapted) hyper-parameters, known to be robust. We only impose the initial mean - randomly drawn within the domain of interest - and covariance of their initial distribution, as well as their population size to match it with the one we use for GENNES, which we set to $N=20$. The starting points for the multiple L-BFGS runs are taken uniformly at random over the domain of interest, and each run stops after a convergence criterion is met. Details on the baseline settings can be found in Appendix \ref{sec::baselines}. Note that we don't compare with derivative-enabled BO methods as their runtime is excessively large for the regimes we consider.

\paragraph{Results and analysis} 
	
    \begin{table}
    \begin{center}
    	\begin{tabular}{c*{4}{c}*4{c}}
        	\toprule
            & \multicolumn{4}{P{0.4\linewidth}}{\textbf{Rastrigin} ($d=10$)} & \multicolumn{4}{P{0.4\linewidth}}{\textbf{Rastrigin} ($d=30$)}   \\
             \T \# evaluations & 1e2 & 1e3 & 1e4 & 1e5 & 1e2 & 1e3 & 1e4 & 1e5  \\
            \hline
            \T \emph{GENNES} & 71.3 & 	41.3 & 	\textbf{4,1} & \textbf{3.9} & 279.0 & 212.1 & 72.3 & \textbf{19.0}  \\
            \emph{LBFGS} & \textbf{24.3}&\textbf{13.3}&6.9&5.4&\textbf{90.4}&\textbf{74.5}&60.7&41.2 \\
            \emph{sNES} &107.2&48.1&4.3&4.2 & 333.5&242.4&40.3&20.8\\
            \emph{aCMA-ES} & 100.9&40.3&6.1&6.0 & 100.9&233.2&\textbf{37.0}&37.0 \\
            \bottomrule
        \end{tabular}
        
        \begin{tabular}{c*{4}{c}|*4{c}}
        	\toprule
            & \multicolumn{4}{P{0.4\linewidth}}{\textbf{Ackley} ($d=10$)} & \multicolumn{4}{P{0.4\linewidth}}{\textbf{Ackley} ($d=30$)} \\
             \T \# evaluations& 1e2 & 1e3 & 1e4 & 1e5 & 1e2 & 1e3 & 1e4 & 1e5\\
             \hline
            \T \emph{GENNES} & \textbf{7.9} &	2.5 & 0.007 & 0.005 & \textbf{9.4} &\textbf{5,4} &	0.007 & 0.006 \\
            \emph{LBFGS} & 11.9&	8.3&5.1&3.2&12.9&11.6&9.0&7.2 \\
            \emph{sNES} &11.4&1.6&	1e-7&5e-10 & 18.7&10.5&9e-05&2e-8\\
            \emph{aCMA-ES} & 10.7&	\textbf{0.2}&\textbf{2e-11}&\textbf{1e-14} & 18.6&	5.9&\textbf{2e-09}&\textbf{1e-14} \\
            \bottomrule
        \end{tabular}
        
        \begin{tabular}{c*{4}{c}|*4{c}}
        	\toprule
            & \multicolumn{4}{P{0.4\linewidth}}{\textbf{Styblinski} ($d=10$)} & \multicolumn{4}{P{0.4\linewidth}}{\textbf{Styblinski} ($d=30$)} \\
             \T \# evaluations& 1e2 & 1e3 & 1e4 & 1e5 & 1e2 & 1e3 & 1e4 & 1e5\\
             \hline
            \T \emph{GENNES} & 206.1&	111.1&\textbf{7.8}&\textbf{5.2} & 2371.3&	587.7&	97.2&\textbf{21.1} \\
            \emph{LBFGS} & \textbf{35.3}&	\textbf{24.0}&9.9&5.6&\textbf{169.0}&\textbf{113.1}&98.2&67.8 \\
            \emph{sNES} &888.6&39.3&11.3&11.1& 19022.0&	647.3&	\textbf{70.6}&68.9\\
            \emph{aCMA-ES} & 657.6&28.9&24.1&23.9 &14213&488.2&131.4&85.4 \\
            \bottomrule
        \end{tabular}
        \begin{tabular}{c*{4}{c}|*4{c}}
        	\toprule
            & \multicolumn{4}{P{0.4\linewidth}}{\textbf{Schwefel} ($d=10$)} & \multicolumn{4}{P{0.4\linewidth}}{\textbf{Schwefel} ($d=30$)} \\
             \T \# evaluations & 1e2 & 1e3 & 1e4 & 1e5 & 1e2 & 1e3 & 1e4 & 1e5\\
             \hline
            \T \emph{GENNES} & 3011.9&	2186.3&\textbf{595.8}&\textbf{533.6}& 10100.6&8381.9&\textbf{1235.4}&\textbf{943.8} \\
            \emph{LBFGS} & \textbf{1463.2}&\textbf{989.7}&718.5&542.8&\textbf{5021.4}&\textbf{4540.6}&3842.6&3331.7\\
            \emph{sNES} &3087.1&	2587.7&685.3&685.3&20807.6&10253.4&2447.8&2429.4\\
            \emph{aCMA-ES} & 3042.3&2301.8&1749.1&1739.2 &13000.4&100021.2&5661.7&5658.7  \\
            \bottomrule
        \end{tabular}
     \end{center}
     \caption{Comparison of best found objective value on Rastrigin, Ackley, Styblinski and Schwefel functions in dimensions $d=10,30$, as a function of number of objective evaluations. Results are averaged over 10 repetitions. Experiments are stopped if all methods have converged or if the maximum budget has been reached.}
    \label{table::exp}
    \end{table}
     
      In Table \ref{table::exp} we report the  difference between the best value found by GENNES, sNES, aCMA-ES and repeated L-BFGS and the value of the objective's minimum as a function of the number of function evaluations. It is noticeable that on the Rastrigin, Styblinski and Schwefel functions, repeated LBFGS is the best algorithm for small number of evaluations, as it greedily exploits the landscape and finds local minima very fast. However, as the number of evaluations grows, the early exploration performed by GENNES pays off and it finds the best minima out of all methods within the given budget. Note that on these three functions, GENNES performances increase compared to other methods as the dimension grows. This is to be expected, as repeated L-BFGS suffers from the curse of dimensionality, and as GENNES leverages the dimension independent oracle complexity of gradient-based methods. In a very highly multimodal function like Ackley, GENNES is better than its baselines for small number of evaluations, and we again see its performances increasing with the dimensionality of the problem. However, it gets stuck near the global optimum when zero order methods manage to find the global optimum very precisely. It is very likely that this phenomenon is caused by the sharp cliff surrounding the global minimum of Ackley's function, making the task harder for gradient-based methods.}
    }
    \subsection{Accelerating Bayesian Optimization}
    {
    	\label{subsec::accbo}
        We here present a possible application of GENNES, which is the efficient minimization of the acquisition function from a BO procedure with GPs. We propose to compare GENNES for optimizing the acquisition function with the repeated L-BFGS method - which is commonly used for this purpose in the BO community. The acquisition function we use is the EI - although results are reproducible for any differentiable acquisition function, while the surrogate is a GP with Automatic Relevance Determination Matern52 kernel. We run comparisons on two tasks: a low-dimensional hyper-parameter optimization task, and a high-dimensional multimodal toy function. Experiments are ran using the library GPyOpt \cite{gpyopt2016}.

        We first apply BO to the optimization of the hyper-parameters of a logistic regression model on the digits dataset from \texttt{scipy}. The four hyper-parameters to optimize are the learning rate, the L2 and L1 regularization coefficients and the number of iterations. Figure \ref{fig::hyperoptdigits} displays the test error as a function of both the number of objective function evaluations and acquisition function queries. The curves are averaged over a 10-fold experiment. It is noticeable that GENNES finds better queries than L-BFGS after maximizing the acquisition function, but also does so in less acquisition queries. We repeat this experience in Figure \ref{fig::alpine1d20} on the toy multimodal function Alpine1 in dimension $d=20$, with similar conclusions. Note that as the problem dimension will grow, the improvement of GENNES over L-BFGS is likely to increase as there would be more local minima. Extensive details about this experimental set-up can be found in Appendix \ref{sec::boexpappendix}.
        
        \begin{figure}
        \centering
        	\begin{subfigure}{0.45\linewidth}
            	\includegraphics[width=\linewidth]{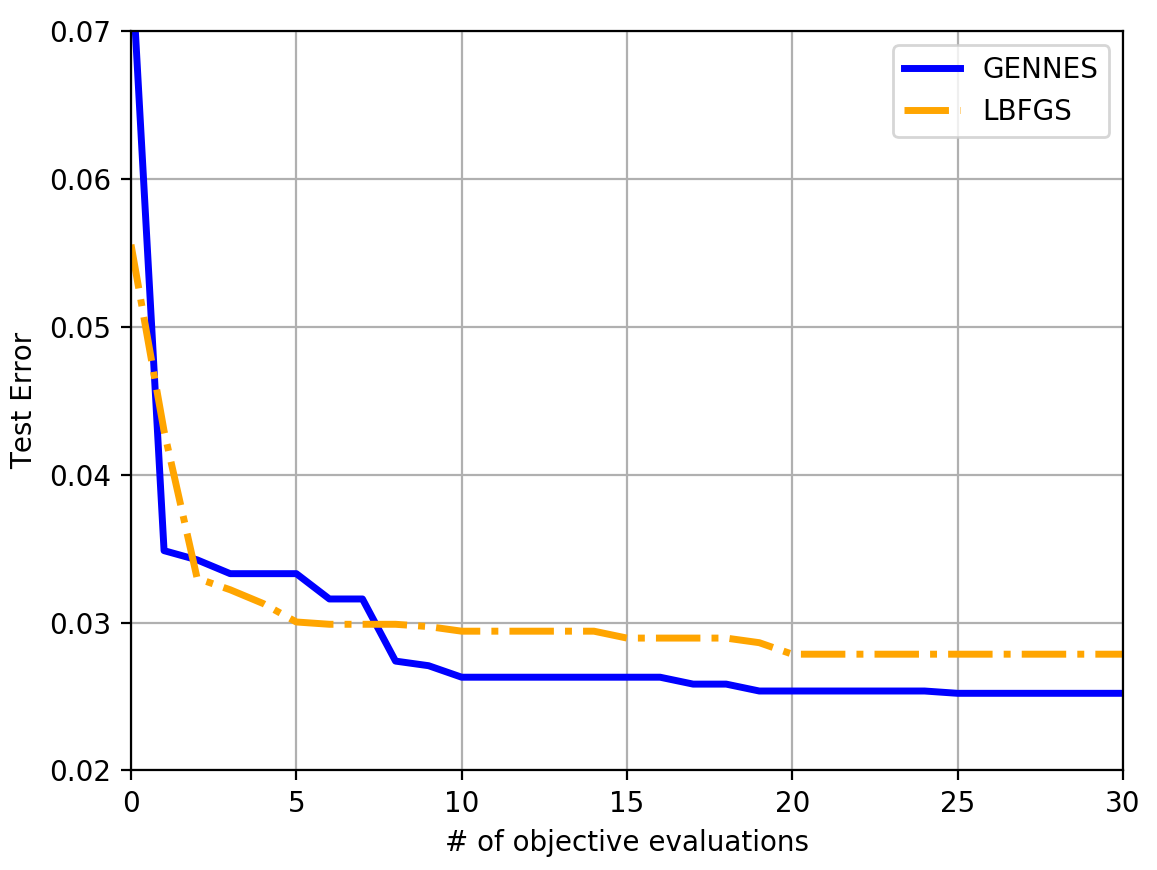}
            \caption{Test error versus number of objective evaluations.}
            \end{subfigure}\hfill
            \begin{subfigure}{0.45\linewidth}
            	\includegraphics[width=\linewidth]{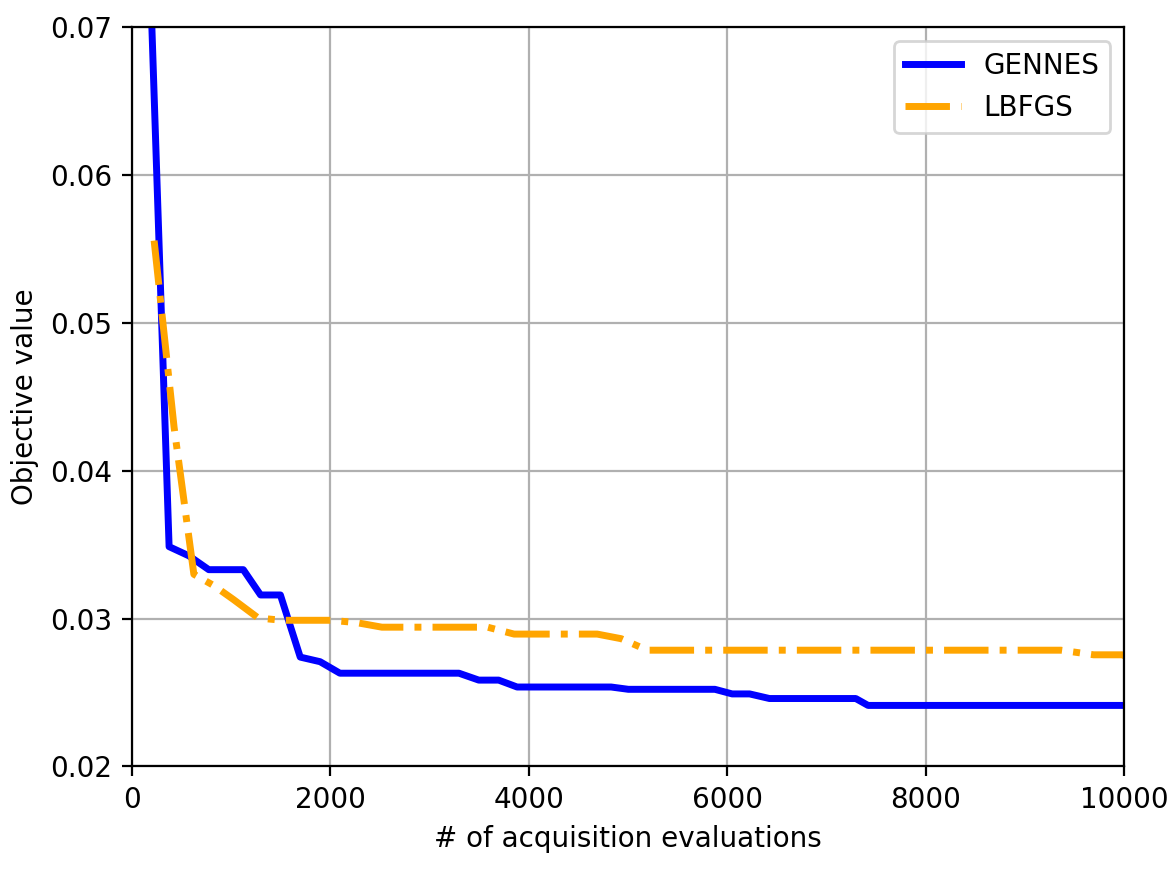}
            \caption{Test error versus number of acquisition function queries.}
            \end{subfigure}
            \caption{Comparison of GENNES and repeated L-BFGS for hyper-parameters optimization via BO on the digits dataset ($d=4$). Results are average over 10 repetitions.}
            \label{fig::hyperoptdigits}
        \end{figure}
        
        \begin{figure}
        \centering
        	\begin{subfigure}{0.45\linewidth}
            	\includegraphics[width=\linewidth]{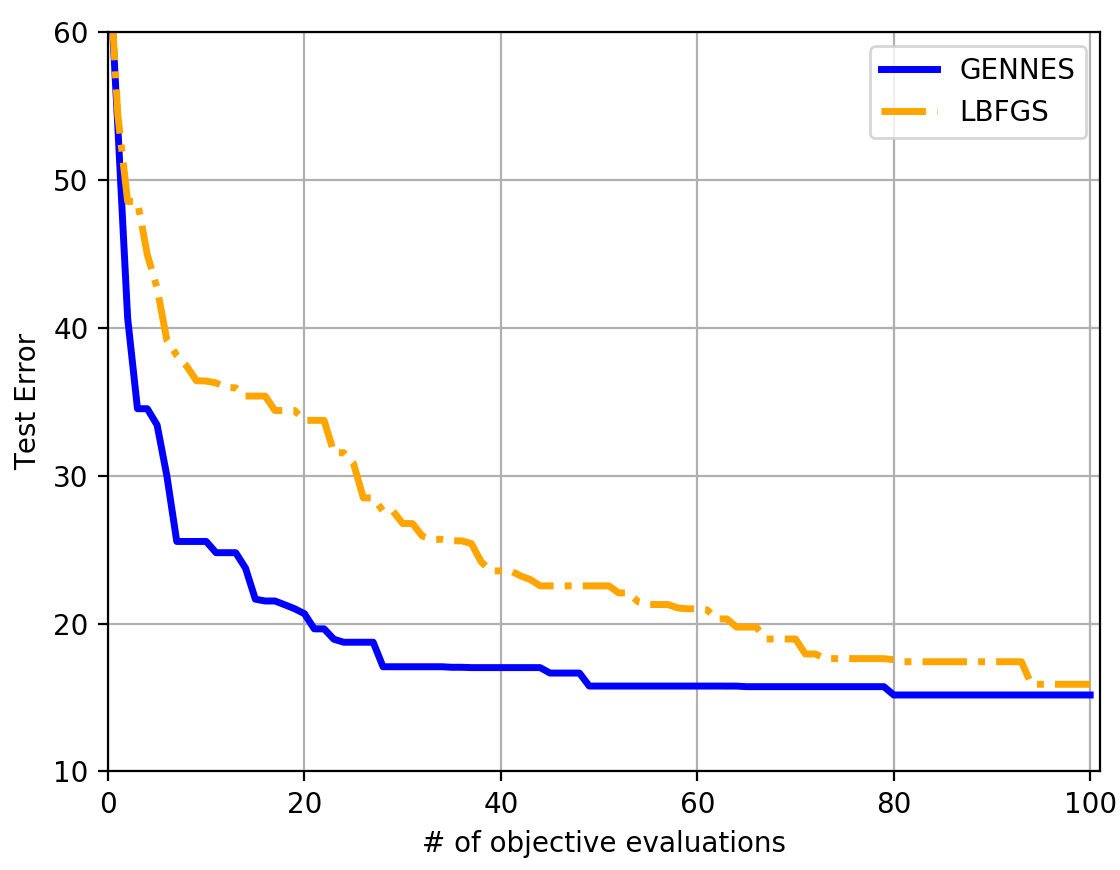}
            \caption{Objective value versus number of objective evaluations. Results are average over 10 repetitions.}
            \end{subfigure}\hfill
            \begin{subfigure}{0.45\linewidth}
            	\includegraphics[width=\linewidth]{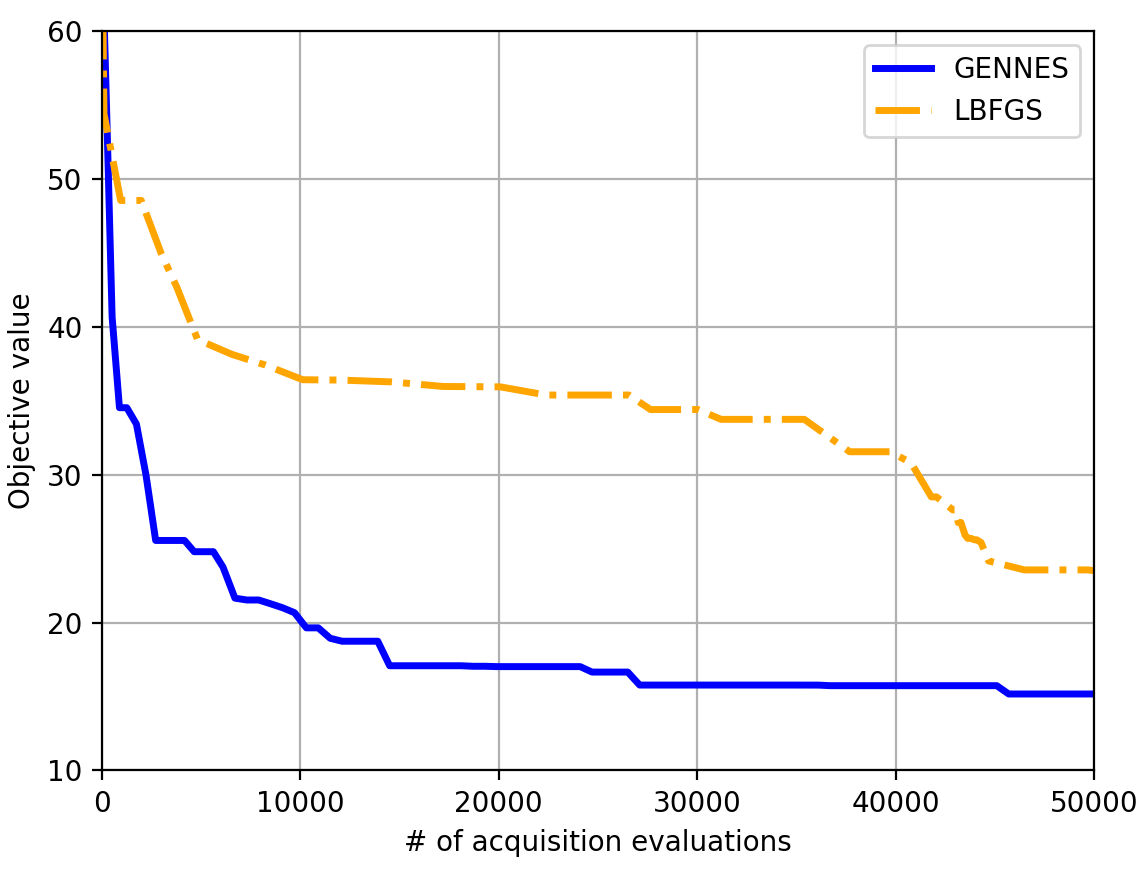}
            \caption{Objective value versus number of acquisition function queries.}
            \end{subfigure}
            \caption{Comparison of GENNES and repeated L-BFGS for global optimization of the Alpine1 function ($d=20$). Results are average over 10 repetitions.}
            \label{fig::alpine1d20}
        \end{figure}
 
    }
}

\section{Conclusion}
{
	\label{sec::conclusion}
	We propose GENNES, a neural generative model to optimize multimodal black-box functions for which gradients are avalaible. We show the merits of our approach on benchmark set of multimodal functions by comparing with state-of-the-art zero order methods and a repeated gradient-based greedy method. We propose to use GENNES to accelerate Bayesian Optimization by efficiently maximizing acquisition functions. We show that we are able to outperform the most popular solution for this task, leading to faster discoveries of the objective global minimum. We believe this to be an important contribution, as Bayesian Optimization is often limited by the cost of evaluation of the acquisition function when the number of query points is large. Finding ways to optimize the acquisition function in less queries is therefore a step forward in scaling Bayesian Optimization.
    
    Other applications of this new method are numerous. In future work, we wish to use recent methods \cite{maclaurin2015gradient} to directly compute hyper-parameters gradients and compare to recent gradient-based Bayesian Optimization methods \cite{wu2017bayesian}. Another promising application of our method is the efficient global optimization of deep neural networks, which we also plan to tackle in future work. 
}

\bibliography{bib}
\bibliographystyle{plain}

\newpage
\appendix
\section{Experimentations details}
{
	\label{sec::experimentsappendix}
	\subsection{Benchmark functions}
	\label{sec::functionformulas}
    Table \ref{table::functionformulas} details the formulas of the benchmark functions from \ref{sec::continuousopt} for a given dimension $d$. All functions have a minimum value $f^*=0$. In our experiment, the objective is translated at every new fold to shift the position of the global minimum.  
    
    \renewcommand{\arraystretch}{1.5}
    \begin{table}[h!]
    	\begin{center}
        	\begin{tabular}{|c|c|c|}
            	\hline
                \textbf{Function} & \textbf{Literal expression} & \textbf{Domain}\\
                \hline
                Rastrigin & $f(x) = 10d + \sum_{i=1}^d\left(x_i^2-10\cos(2\pi x_i)\right)$ & $[-3,3]^d$\\
                \hline
                Ackley & $f(x)=-20\exp{\left(-0.2\sqrt{\frac{1}{d}\sum_{i=1}^d x_i^2}\right)}-\exp{\left(\frac{1}{d}\sum_{i=1}^d \cos {x_i}\right)} +20 + e$ & $[-10,10]^d$\\
                \hline
                Styblinski & $\frac{1}{2}\sum_{i=1}^d x_i^4-16x_i^2+5x_i$ & $[-10,10]^d$\\
                \hline
                Schwefel & $418.9928\cdot d - \sum_{i=1}^d x_i\sin{\left(\sqrt{x_i}\right)}$& $[-500,500]^d$\\
                \hline
                Alpine1 & $f(x)=\left| \sum_{i=1}^d x_i\sin{(x_i)}+0.1x_i\right|$ & $[-10,10]^d$ \\
                \hline
            \end{tabular}
        \end{center}
        \caption{Benchmark functions literal expressions.}
        \label{table::functionformulas}
       	\end{table}
    
    \subsection{Baselines settings}
    { 
    	\label{sec::baselines}
    	For every experiment, we set the population size of aCMA-ES, sNES and GENNES to $N=20$. At every new fold, we randomly impose the mean of the initial distribution of aCMA-ES and sNES to lie within the objective domain. The standard deviation is set to properly cover the whole domain. All other hyper-parameters for both of these methods are left untouched, as their author provide robust adapted methods to set them (see \cite{hansen2016cma,schaul2012benchmarking}).
        
        For GENNES, a new seed is used at every new fold for the initialization of the weights and biases of the generator. The network is composed of $6$ fully connected layers with leaky ReLU activations (with $\alpha=0.2$) and a last fully connected layer with tanh activation. The variance of the weights of this last layer are chosen according to Appendix \ref{sec::safeinit} so that the output distribution has proper adapted variance that covers the domain. The noise annealing schedule is chosen according to Appendix \ref{sec::noiseschedule}, with coefficient $\alpha = 0.99$ for all the different experiments.
    }
    
    \subsection{Bayesian Optimization experiments}
    {
    	\label{sec::boexpappendix}
        
        The first experiment we run tackles the optimization of the hyper-parameters of a logistic regression modal applied on the digits dataset, made up of 1797 8x8 images, representing a hand-written digit. The train/test split ratio is set to $r=0.15$. The optimized hyper-parameters are the learning rate on a log-scale in $[10^{-5},1]$, the L2 and L1 regularization parameters on a log-scale in $[10^{-5},1]$ and the number of iterations between $[5,50]$. The generator used for GENNES is made up of 6 fully connected layers with leaky ReLU activations for the first 5 ones and with hyperbolic tangent for the last. The exponential scheduling factor is set to $\alpha=0.99$ for all the experiments, and the population size to $N=10$. We run GENNES for 20 iterations. At every iteration, we sample 20 starting point for L-BFGS, which is stopped when the minimum coordinate of the projected gradient has a magnitude lower than $10^{-5}$. The initial GP is fit with $20$ points. 
        
        The second experiment deals with the global optimization of the Alpine1 function on $[-10,10]^d$ in dimension $d=20$. The generator of GENNES is left untouched, except for the maximum number of iterations which is brought up to 30. We use 100 initial points for L-BFGS at every iteration, and initialize the GP with 100 points. 
    }
}

\section{Guaranteeing a safe initialization for the generator}
{ 
	\label{sec::safeinit}
	We hereinafter describe the reasoning we follow to set the initial distribution of the input noise $u$, which dimension we note $p$. Let us consider that $u$ is sampled along a multivariate uniform distribution, which first and second moments we respectively note $\mu$ and $\nu^2$. The weights $w$ of the first layer are sampled according to a centered multivariate normal distribution of variance $\sigma_0^2$. The activation $z$ of a neuron belonging to the first hidden layer is obtained via: 
    $$
    	z_i = \sum_{j=1}^p w_{i,j}u_j
    $$
    as the initial biases are set to $0$. Therefore we can write the mean and variance of its related random variable (r.v) $Z$ as:
    \begin{equation}
    	\begin{aligned}
    		\mathbb{E}\left[Z\right] &=  0\\
            \mathbb{V}\left[Z\right] &= p\sigma_0^2\left(\nu^2+\mu^2\right)
            \label{eq::Amoments}
        \end{aligned}
    \end{equation}
    assuming of course that the random variables representing the weights and the noise are independent. We approximate $Z$'s distribution as a normal r.v with moments matching \eqref{eq::Amoments}. We want to approximate the moments of the distribution obtained after feeding $z_i$ to a ReLU activation (we note $a_i$ the resulting scalar and $A$ its related r.v). Easy computations show that $Z$ being centered, we have that:
    $$
    	\begin{aligned}
        	\mathbb{E}\left[A\right] &= \sqrt{\mathbb{V}\left[Z\right]/2\pi}\\
           							 &= \sigma_0\sqrt{\frac{p}{2\pi}(\nu^2+\mu^2)}
        \end{aligned}
    $$
    and
    $$
    	\begin{aligned}
        	\mathbb{V}\left[A\right] &= \mathbb{V}\left[Z\right]/2 - \mathbb{E}\left[A\right]^2 \\
            						 &= \frac{p\sigma_0^2}{2}\left(\nu^2+\mu^2\right)(1-\frac{1}{\pi})
        \end{aligned}
    $$
    
    Assuming for the sake of simplicity that $u$ is centered we can simplify that last expression as:
    $$
    	\mathbb{V}\left[A\right] = \frac{p\sigma_0^2\nu^2}{2}(1-\frac{1}{\pi})
    $$
    If all $n$ the following layers now have the same height $h$, repeating this process indicates that the variance of the activations of the last hidden layer before the tanh activation is given by:
    $$
    	\begin{aligned}
    		\mathbb{V}\left[A_{n}\right] &= p\sigma_0^2\nu^2\left(\frac{h\sigma^2}{2}(1-\frac{1}{\pi})\right)^{n}\\
            							& \simeq p\sigma_0\nu^2\left(0.3h\sigma^2\right)^{n}
        \end{aligned}
    $$
    with $\sigma$ being the variance of the weights in the following layers.
    Now, the weights of the ReLU layers being initialized with Glorot initialization, we have the $\sigma_0^2=1/p$ for the first layer, and $\sigma^2=1/h$ for the rest. Now denoting $\lambda^2$ the variance of the distribution used to generate the weights of the very last layer and $\beta^2$ the variance we wish to impose to the output distribution, and approximating the hyperbolic tangent by the identity (the distribution of the activation is centered) we are left with the relation (after straight forward simplifications):
    $$
    	\lambda^2 h \nu^2(0.3)^{n} \simeq \beta^2
    $$
    Finally, we are left with the following equality:
    \begin{equation}
    	\lambda^2\nu^2 = \frac{\beta^2}{h(0.3)^{n}}
        \label{eq::finallayervariance}
    \end{equation}
    which dictates how to tune the product $\lambda^2\nu^2$ as a function of the generator's depth and the desired variance for the output distribution (which is centered around $0_n$ as the weights of the last layer are sampled according to a centered distribution. Even though it requires some questionable approximations, we found this formula to work very well in practice for small networks (typically with $n\leq 10$ and $h\leq 10^3$. The mean of the initial distribution can easily be shifted as required by the optimization problem at hand by adding bias on the output layer. 
    
    Notice that the dimension $p$ of the noise does not appear in \eqref{eq::finallayervariance}. In practice, it is important to have $p$ close to $d$ as an excessively small value results in a weirdly shaped support for the output distribution, which hinders the optimization as the resulting distribution of gradients might mislead the generator. Indeed, the input dimension of the noise impacts the number of possibly uncorrelated output dimensions. For instance with input of size one all the outputs dimensions would be correlated.

}

\section{Setting the noise annealing schedule}
{ 
	\label{sec::noiseschedule}
    The setting of the noise annealing schedule is crucial, as an early collapse of the output distribution could lead to a sequence of pure gradient-descent like updates which could mean that GENNES would potentially miss the global optimum. On the other hand, if high-precision for the position of the global optimum is required, the annealing should decrease quickly the support of the noise once the global minimum has been located. In practice, we found that an exponential decrease rule such as:
    \begin{equation}
    	u_t = \alpha^t u_0
    \end{equation}
    where $u_0$ is the initial noise (with initial magnitude) and $\alpha<1$  is the parameter dictating the evolution of the noise support. If it is likely that the objective has few local minima and a convex shape near them, then we recommend setting $\alpha$ to a small value, \emph{i.e} like $\alpha = 0.9$. On the other hand, if the objective is highly multimodal, we recommend setting it to a value closer to 1, such as $\alpha=0.999$. 
}

\end{document}